\documentclass[journal,twoside,web]{ieeecolor}
\usepackage{jsen}
\usepackage{cite}
\usepackage{amsmath,amssymb,amsfonts}
\usepackage{algorithmic}
\usepackage{graphicx}
\usepackage{textcomp}
\usepackage{wrapfig}
\usepackage{placeins}
\usepackage{balance}
\usepackage{tabularx}
\usepackage{array}
\usepackage{multirow}

\def\BibTeX{{\rm B\kern-.05em{\sc i\kern-.025em b}\kern-.08em
    T\kern-.1667em\lower.7ex\hbox{E}\kern-.125emX}}
\definecolor{abstractbg}{rgb}{0.89804,0.94510,0.83137}

\usepackage[hidelinks]{hyperref}
\makeatletter
\def\ps@titlepagestyle{\def\@oddfoot{}\def\@evenfoot{}%
  \def\@oddhead{\begin{tabular}{@{}p{43pc}@{}}\\[-24pt]\vbox{\hsize43pc\scriptsize\textsf\leftmark \hfill \textsf\thepage\hbox{}}\\[-19pt]
     \vbox{\color{subsectioncolor}\hrule height1pt width43pc depth0pt} \end{tabular}}%
  \def\@evenhead{\@oddhead}%
  \footskip 16pt%
  \def\@oddfoot{\hss\normalfont\footnotesize\parbox[t]{43pc}{\centering\color{red}\bfseries
     This work has been submitted to the IEEE for possible publication. Copyright may be transferred without notice, after which this version may no longer be accessible.}\hss}%
  \def\@evenfoot{\@oddfoot}}
\long\def\IEEEtitleabstractindextext#1{\def\@IEEEtitleabstractindextext{\parbox{1\textwidth}{{\color{subsectioncolor}\hrule}\bigskip\par#1\bigskip\par{\color{subsectioncolor}\hrule}\smallskip\par
  \centering{\color{red}\large\bfseries Supplementary video: A video demonstrating real-time operation of the complete visuo-inertial inference pipeline is available at \url{https://doi.org/10.6084/m9.figshare.34442622}\par}\vspace*{-\videonotepull}}}}
\newlength{\videonotepull}
\makeatother
\begin{document}
\title{Tread: A Visuo-Inertial Framework for Terrain and Locomotion Context-Aware Gait-Phase and Speed Estimation in Wearable Robotics}

\author{Debadrata Sarkar, Akash Bachhar, Aman Arora \textsuperscript{*}
\thanks{This work was fully supported by the CSIR-National Laboratory Scheme under the ULIP [MLP002613] and the CSIR Mission Mode Project [NCP010303].}
\thanks{This work involved human subjects. Approval of all ethical and experimental procedures was granted by the 7th sitting of the Institutional Ethical Board of CSIR-CMERI, Durgapur, India, on 27 February 2024.}
\thanks{The authors are with the Human Centred Robotics and Cybernetics Group, CSIR-Central Mechanical Engineering Research Institute, Durgapur 713209, India. }
\thanks{ \textsuperscript{*}Corresponding author: Dr Aman Arora (e-mail: aman.cmeri@nic.in)}}

\IEEEtitleabstractindextext{%
\fcolorbox{abstractbg}{abstractbg}{%
\begin{minipage}{\textwidth}%
\setlength{\intextsep}{0pt}
\begin{wrapfigure}[11]{r}{3.8in}%
\includegraphics[width=3.8in]{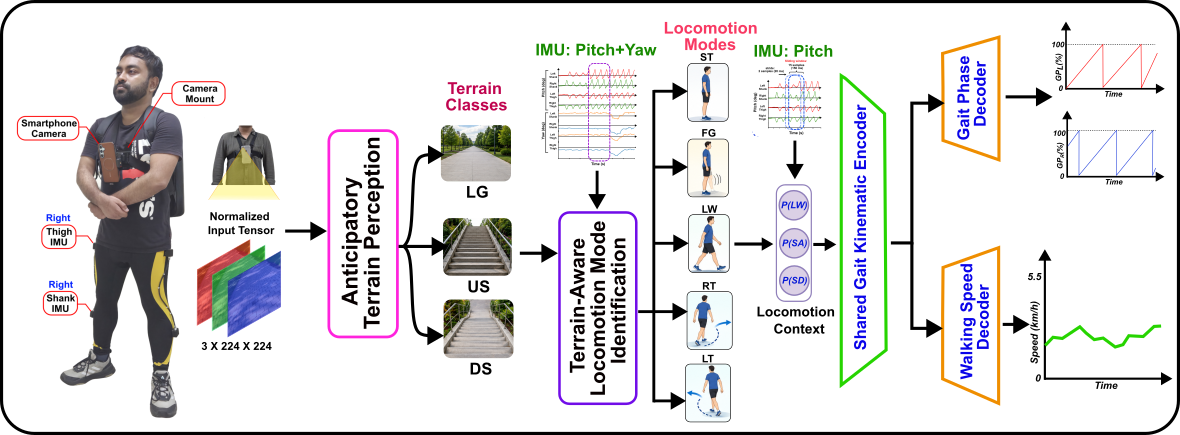}%
\end{wrapfigure}%
\begin{abstract}

A wearable assistive robot must continuously estimate terrain, locomotion mode,
gait phase, and walking speed. These quantities are commonly addressed by
separate, terrain-specific models, with datasets rarely collected under a
common protocol. This paper presents a visuo-inertial framework that connects
these estimates through a causal pipeline using a chest-mounted camera and four
lower-limb inertial measurement units (IMUs). Anticipatory Terrain Perception
(ATP) classifies incoming terrain with 96.3\% accuracy and anticipates terrain
transitions by 1.46~s on average. Terrain-Aware Locomotion Mode Identification
(TALMI) achieves 93.7\% and 98.8\% accuracy for its level-ground and stairs
experts respectively, with mean transition latencies being below 200~ms. The relevant TALMI
posterior is retained as a Probabilistic Locomotion Context (PLC) vector and
provided to a shared Gait Kinematic Encoder (GKE), which supports 50-Hz
bilateral gait-phase estimation and 2-Hz walking-speed estimation. Leave-one-
subject-out evaluation yields gait-phase RMSEs of 1.75\%, 2.97\%, and 4.55\%
for level walking, stair ascent, and stair descent respectively while estimating speed over 3.5-5.5 km/h with a RMSE of 0.218 $\pm$ 0.072 km/h. All inertial sensing based models use sensor-fused Euler angles as input rather than raw accelerometer and gyroscope signals, reducing dependence on sensor mounting variation.

\end{abstract}

\end{minipage}}}

\maketitle

\section{Introduction}
\label{sec:intro}

\IEEEPARstart{P}{owered} lower-limb exoskeletons, exosuits, and active prostheses
require timely information about the user's locomotion for adapting assistance to gait
phase, walking speed, and terrain. Errors in timing or locomotion mode can reduce
assistance quality and compromise stability \cite{kang2021, molinaro2024}.
Recent reviews identify intention recognition and synchronization with the user as
key requirements for wearable-robot autonomy \cite{vanderkooij2025}. Accordingly,
a practical perception layer should estimate four quantities continuously: upcoming
terrain, locomotion mode, gait phase, and walking speed. Although these quantities
are often estimated separately, their integration into a single body-worn pipeline
remains less explored.

Vision provides anticipatory information about terrain, with egocentric datasets
supporting terrain recognition and its use for terrain-aware assistance \cite{exonet, laschowski2022, tricomi2023, tricomi2025, exosense2025}.
IMUs mounted on the lower limbs provide low-power, illumination-independent
measurements for locomotion-mode, gait-phase, and speed estimation
\cite{ma2025, ji2025}. Gait phase estimation has been addressed using both discrete and continuous formulations
\cite{mazon2022, wu2022, tang2024}, while walking speed is commonly estimated
with a dedicated regressor \cite{maldonado2025}. While plantar Force Sensitive Resistor (FSR)s are also widely used to derive ground truth \cite{kang2021} of gait-events, they are rarely used in deployment.

\begin{figure}[!t]
\centering
\includegraphics[scale=1.0]{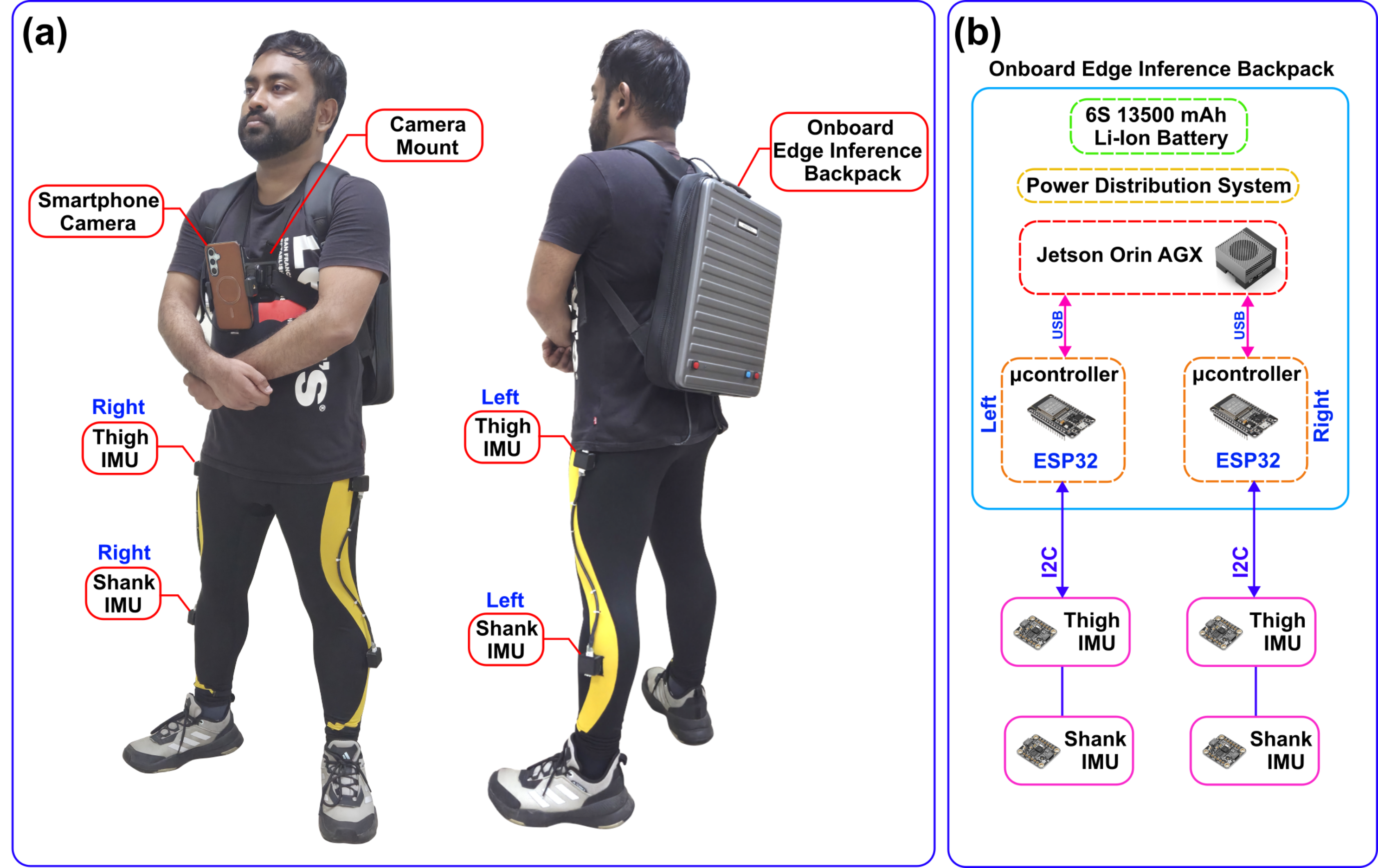}
\caption{Wearable visuo-inertial platform: front and rear views of the instrumented participant.}
\label{fig1}
\end{figure}

Several limitations in state-of-the-art literature motivate the framework proposed through this work. Existing systems commonly treat terrain, locomotion mode, gait phase, and speed as independent tasks, while coupled approaches may rely on hard switching between mode-specific estimators, which produces undesirable estimations near terrain/locomotion mode transitions \cite{qian2022}. Gait-phase estimation accuracy has also been reported to degrade during terrain changes \cite{ji2025}. In addition, a unified system requires synchronized
vision, multi-segment inertial measurements, gait-phase ground truth, and speed
labels, although these are most conveniently collected under different protocols.
Preparatory leg movements pose further challenge because stationary fidgeting can
otherwise be mistaken for walking. Finally, raw accelerometer and gyroscope signals can be sensitive to sensor mounting and re-donning variability \cite{zhu2023}.
Table~\ref{tab_comparison} compares the proposed framework with representative
studies.

\begin{table*}[!t]
\caption{COMPARISON WITH REPRESENTATIVE WORKS}
\label{tab_comparison}
\centering
\scriptsize
\setlength{\tabcolsep}{2.5pt}
\renewcommand{\arraystretch}{1.05}

\begin{tabularx}{\textwidth}{|
>{\raggedright\arraybackslash}p{1.65cm}|
>{\raggedright\arraybackslash}p{1.65cm}|
>{\raggedright\arraybackslash}p{1.35cm}|
>{\raggedright\arraybackslash}p{1.25cm}|
>{\raggedright\arraybackslash}p{1.75cm}|
>{\raggedright\arraybackslash}p{1.25cm}|
>{\raggedright\arraybackslash}p{1.25cm}|
>{\raggedright\arraybackslash}p{2.5cm}|
>{\raggedright\arraybackslash}X|}
\hline

\textbf{Work (year)} &
\textbf{Sensing} &
\textbf{Inertial signals} &
\textbf{Terrain} &
\textbf{Locomotion modes} &
\textbf{Gait phase} &
\textbf{Speed} &
\textbf{Context coupling} &
\textbf{Indicative result} \\
\hline

Laschowski \textit{et al.} \cite{laschowski2022} (2022) &
RGB vision &
-- &
12 classes &
-- &
-- &
-- &
Perception only &
Best 12-class test accuracy: 73.2\% \\
\hline

Tricomi \textit{et al.} \cite{tricomi2023} (2023) &
RGB vision + 2 IMU &
acc, gyro &
LG, SA, SD &
-- &
Continuous &
-- &
Deterministic switch &
Per-class accuracy $>$85\% \\
\hline

Tricomi \textit{et al.} \cite{tricomi2025} (2025) &
RGB vision (geometric) &
acc, gyro &
LG, stairs &
-- &
-- &
-- &
Geometric context &
Overall terrain-detection accuracy: 93.0 $\pm$ 1.1\% \\
\hline

Qian \textit{et al.} \cite{qian2022} (2022) &
Depth vision + IMU &
acc, gyro &
LG, SA, SD &
LG, SA, SD &
Continuous &
-- &
Predictive switch &
Mode accuracy: $>$98\% \\
\hline

Ma \textit{et al.} \cite{ma2025} (2025) &
IMU only (6) &
acc, gyro &
-- &
5 steady + transitions &
-- &
-- &
None (majority filter) &
Steady-mode accuracy: 99.58\% \\
\hline

Kang \textit{et al.} \cite{kang2021} (2021) &
IMU + hip encoders &
acc, gyro, joint angle &
-- &
-- &
Continuous &
-- &
None (mode-agnostic) &
Gait Phase RMSE: 5.04$\pm$0.79\% \\
\hline

Mazon \textit{et al.} \cite{mazon2022} (2022) &
IMU only (1--2) &
acc, gyro &
-- &
ST, LW, RA, RD, SA, SD (+sit) &
Discrete &
-- &
Joint multi-level (modes+trans.+phases) &
F1: 0.89--0.91 (10 subj., locomotion modes+ Gait phases) \\
\hline

Maldonado \textit{et al.} \cite{maldonado2025} (2025) &
IMU (prosthesis) &
acc, gyro &
-- &
-- &
-- &
Speed (m/s) &
None &
Speed MAE: 0.074 m/s \\
\hline

\textbf{This work} &
\textbf{RGB vision + 4 IMU} &
\textbf{pitch, yaw (Euler)} &
\textbf{LG, US, DS} &
\textbf{ST, FG, LW, RT, LT, SA, SD} &
\textbf{Continuous, bilateral} &
\textbf{Yes (km/h)} &
\textbf{Probabilistic (PLC)} &
\textbf{terrain accuracy: 96.3\% (1.46s anticipation horizon); locomotion mode acuuracy: 93.7\%-98.8\%; gait phase RMSE: 1.75–4.55\%; speed RMSE: 0.218 ± 0.072 km/h} \\
\hline

\end{tabularx}

\vspace{2pt}

\raggedright
\scriptsize
Reported values originate from different sensor sets, participant groups, and protocols, and are indicative rather than directly comparable.
LG: level ground; US/SA: upstairs/stair ascent; DS/SD: downstairs/stair descent; ST: standing; FG: fidgeting; LW: level walking; RT/LT: right/left turn; RA/RD: ramp ascent/descent; acc: accelerometer; gyro: gyroscope; LOSO: leave-one-subject-out.

\end{table*}

The main contribution of this work is a connected visuo-inertial pipeline that
produces terrain, locomotion mode, bilateral gait phase, and walking speed estimation using
a unified sensing suite deployed for onboard edge inference. Anticipatory Terrain Perception (ATP)
uses vision to estimate the upcoming terrain, while Terrain-Aware Locomotion Mode
Identification (TALMI) estimates the locomotion being executed. Rather than
passing only a hard locomotion label downstream, the full TALMI softmax posterior
is retained as a Probabilistic Locomotion Context (PLC) vector. Gait phase and
walking speed are then decoded from a shared Gait Kinematic Encoder (GKE) through
lightweight task-specific heads. A dedicated fidgeting class separates stationary
leg movements from walking and a terrain-conditioned hierarchy allows datasets
collected under different protocols to be combined within one estimation chain.

All inertial models use pitch and yaw Euler angles obtained from onboard sensor
fusion rather than raw accelerometer and gyroscope channels. This representation
is gravity-referenced and expresses limb orientation in an anatomically meaningful
frame, reducing dependence on sensor-frame rotation and small mounting variations
\cite{zhu2023}, that otherwise make the reproducability and scalability difficult when used in a new wearable hardware. The experiments therefore evaluate whether this compact
representation, combined with terrain and locomotion context, is sufficient for
terrain-aware mode identification, continuous bilateral gait-phase estimation, and
walking-speed regression.

\section{Hardware and Data Collection}
\label{sec:hardware}

\subsection{Instrumentation}

The wearable platform is shown in Fig.~\ref{fig1}. A smartphone RGB camera is
mounted at sternum height and tilted 10--15$^\circ$ from vertical to capture
upcoming terrain while retaining the near-field ground. Four IMUs (BNO085, Adafruit Industries) are mounted
laterally on the right thigh, right shank, left thigh, and left shank respectively. Each
bilateral pair is interfaced to an ESP32 over I2C and both microcontrollers
stream data via USB to a Jetson AGX Orin (NVIDIA Corporation). The Jetson, microcontrollers,
and a 6S 13,500-mAh Li-ion battery with its power-distribution system are carried
in a backpack, enabling fully onboard sensing and inference. Data acquisition,
preprocessing and edge inference use ROS~2 and micro-ROS, providing a common
time-stamped middleware layer. During data collection, sensorized insoles with an
FSR beneath each heel provided heel-contact events for gait-phase annotation. These were used only as ground truth gait event markers and are excluded from the deployed system.

\subsection{Participants}

Data were collected from 10 male participants (mean age: 28.7
$\pm$ 4.6 years; mean height: 171.2 $\pm$ 7.4 cm; 8
right-foot dominant), all of whom provided informed consent. Two additional male
participants, excluded from training, validation, and test sets, were used only
for unseen-user demonstrations (ages 29 and 28
years; heights 173 and 169 cm).

\subsection{Acquisition Protocols}

For terrain perception, participants walked at self-selected speeds through
indoor and outdoor environments containing level ground, stair ascent and descent.
Images were recorded at $720\times1280$ pixels and 30 frames per second. For inertial data,
participants followed routes containing level ground, stairs and turns while an
observer followed behind and recorded a second-person video. Participants also stopped, turned and
fidgeted naturally in addition to when prompted to do so, to avoid over-representation of steady walking. Turns were annotated for heading changes
of approximately 60--90$^\circ$. The video was used for inertial-mode annotation,
with heel-contact FSR traces providing additional temporal support. Approximately 20 minutes of overground inertial data were collected per participant for locomotion-mode annotation, in addition to the treadmill sessions for annotated walking speed described below..

For each limb, a gait cycle was defined between successive ipsilateral heel
contacts and phase was linearly interpolated from 0--100\% within each cycle.
Speed-labeled data were collected on a treadmill at 3.5, 4.0, 4.5, 5.0, and
5.5 km/h using three sessions per speed, where each session was approximately of 2 minutes. Only four participants could maintain a continuous 5.5 km/h walk comfortably on treadmill. The treadmill and overground protocols were intentionally
separate. The former provided controlled speed labels, whereas the latter
provided terrain and turning data without direct speed labels. Their integration
within the proposed framework is described in Sec.~\ref{sec:ctx}.

\section{Anticipatory Terrain Perception}
\label{sec:atp}

\begin{figure*}[!t]
\centering
\includegraphics[scale=1.0]{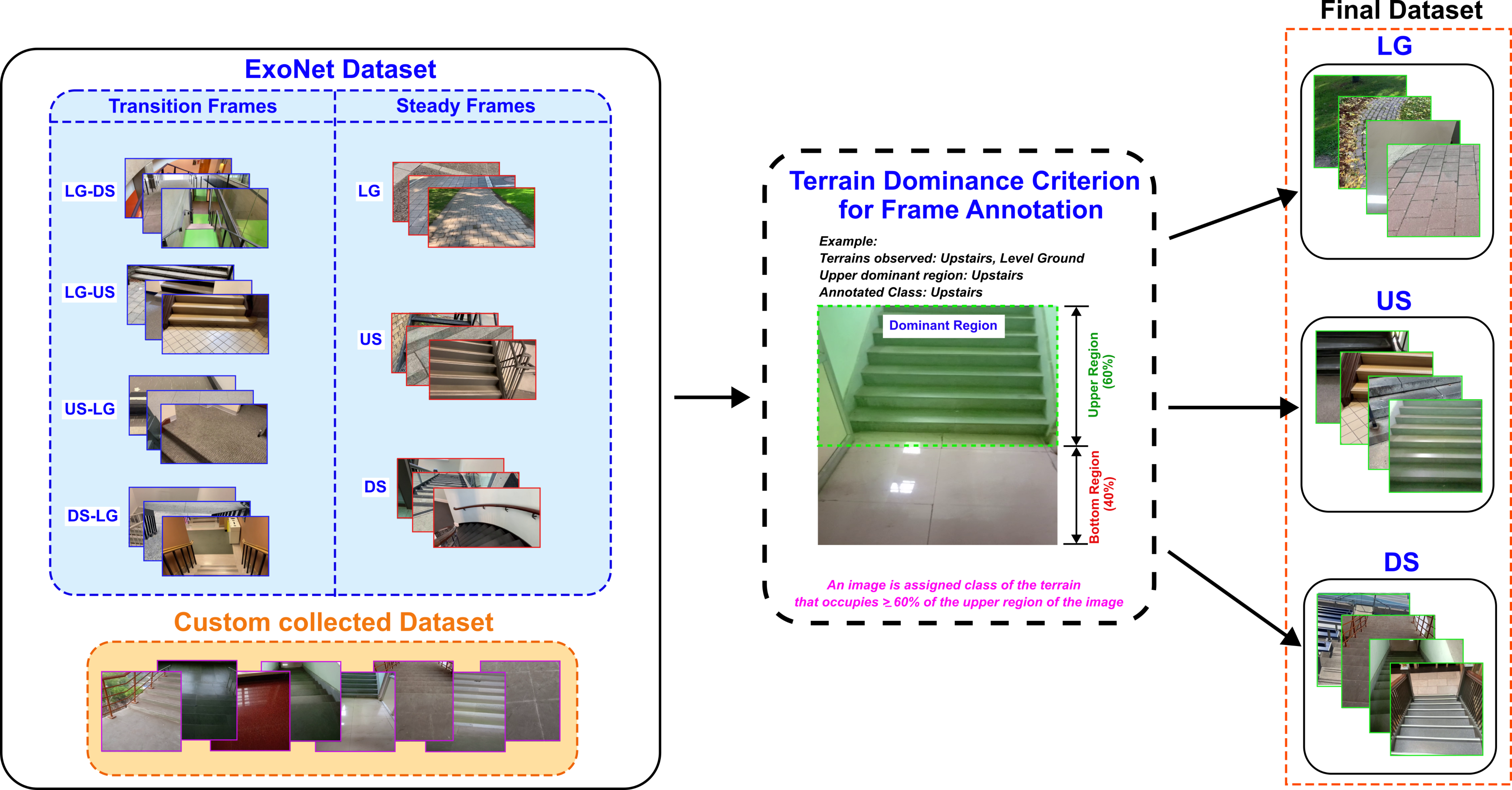}
\caption{(a) ATP dataset construction from ExoNet and custom frames using the terrain dominance criterion and three terrain classes. (b) Power and data architecture of the onboard edge-inference backpack.}
\label{fig2}
\end{figure*}

A terrain classifier that responds only after a new surface has been treaded over cannot
provide advance information for control. ATP therefore classifies the terrain
being approached, providing look-ahead information for downstream estimation as well as several supporting functions utilized during deployment, such as gain scheduling, hardware initialization or actuator pre-charging.

The ATP corpus (Fig.~\ref{fig2}) combines the public ExoNet dataset \cite{exonet},
which contributes 54.48\% of the frames, with a custom recorded chest-camera dataset
(45.52\%). Both datasets use the same annotation rule; the upper 60\% of each
frame is considered the dominant region and the frame is assigned the terrain
class which occupies at least the upper 60\% region. This associates the image with the
incoming terrain observed ahead of the user rather than the terrain directly
underfoot, while providing a deterministic label for mixed-terrain frames.

\begin{figure*}[!t]
\centering
\includegraphics[scale=1.00]{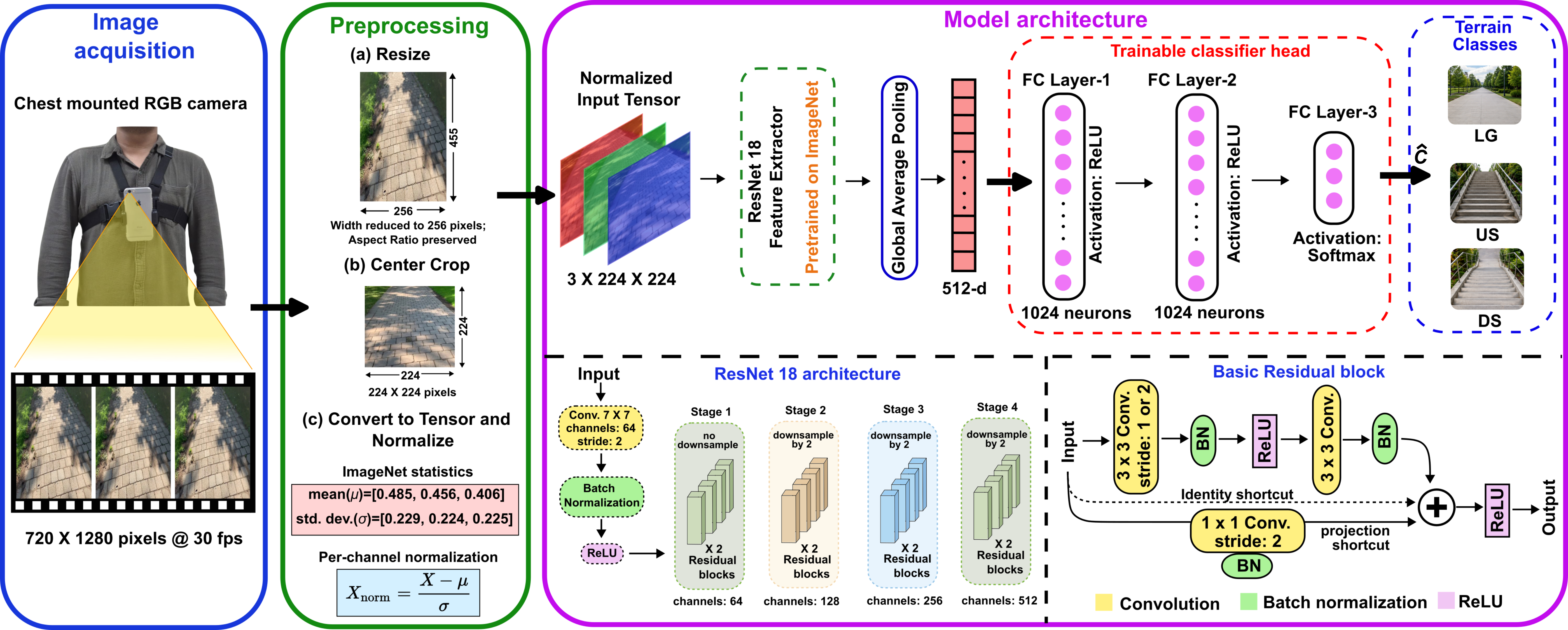}
\caption{ATP pipeline using a frozen ResNet-18 feature extractor and trainable
three-layer classifier head.}
\label{fig3}
\end{figure*}

The annotated frames are divided into training, validation, and test sets using
a 70\%/20\%/10\% split, performed at the sequence level so that no frames from the same video appear in more than one split. Images are resized with the shorter side set to 256
pixels, center-cropped to $224\times224$ pixels and normalized using ImageNet
statistics. ATP uses an ImageNet-pretrained ResNet-18 backbone \cite{resnet} as a
frozen feature extractor(Fig. \ref{fig3}). Global average pooling yields
$\mathbf{z}\in\mathbb{R}^{512}$, followed by a trainable MLP with 1024, 1024 and
3 units. While ReLU activations are used in the first two layers, the final softmax layer produces
the terrain posterior $\mathbf{q}\in\mathbb{R}^{3}$; the predicted terrain being
$\hat{c}=\arg\max_j q_j$. The classifier head is trained for 50 epochs using
Adam optimizer, a learning rate of $10^{-3}$, step decay of 0.1 every 7 epochs, batch size
of 32 and cross-entropy loss function. The ATP is designed to infer the oncoming terrain at 30 Hz.

\begin{table}[!t]
\caption{ATP CLASSIFICATION REPORT ON THE HELD-OUT TEST SET}
\label{tab_atp}
\centering
\setlength{\tabcolsep}{5pt}
\begin{tabular}{|c|c|c|c|}
\hline
\textbf{Class} & \textbf{Precision} & \textbf{Recall} & \textbf{F1-Score} \\
\hline
LG & 0.945 & 0.953 & 0.949 \\
\hline
US & 0.967 & 0.973 & 0.970 \\
\hline
DS & 0.978 & 0.966 & 0.972 \\
\hline
\textbf{Accuracy} & \multicolumn{3}{c|}{0.963} \\
\hline
\end{tabular}
\end{table}

\begin{figure}[!t]
\centering
\includegraphics[scale=1.0]{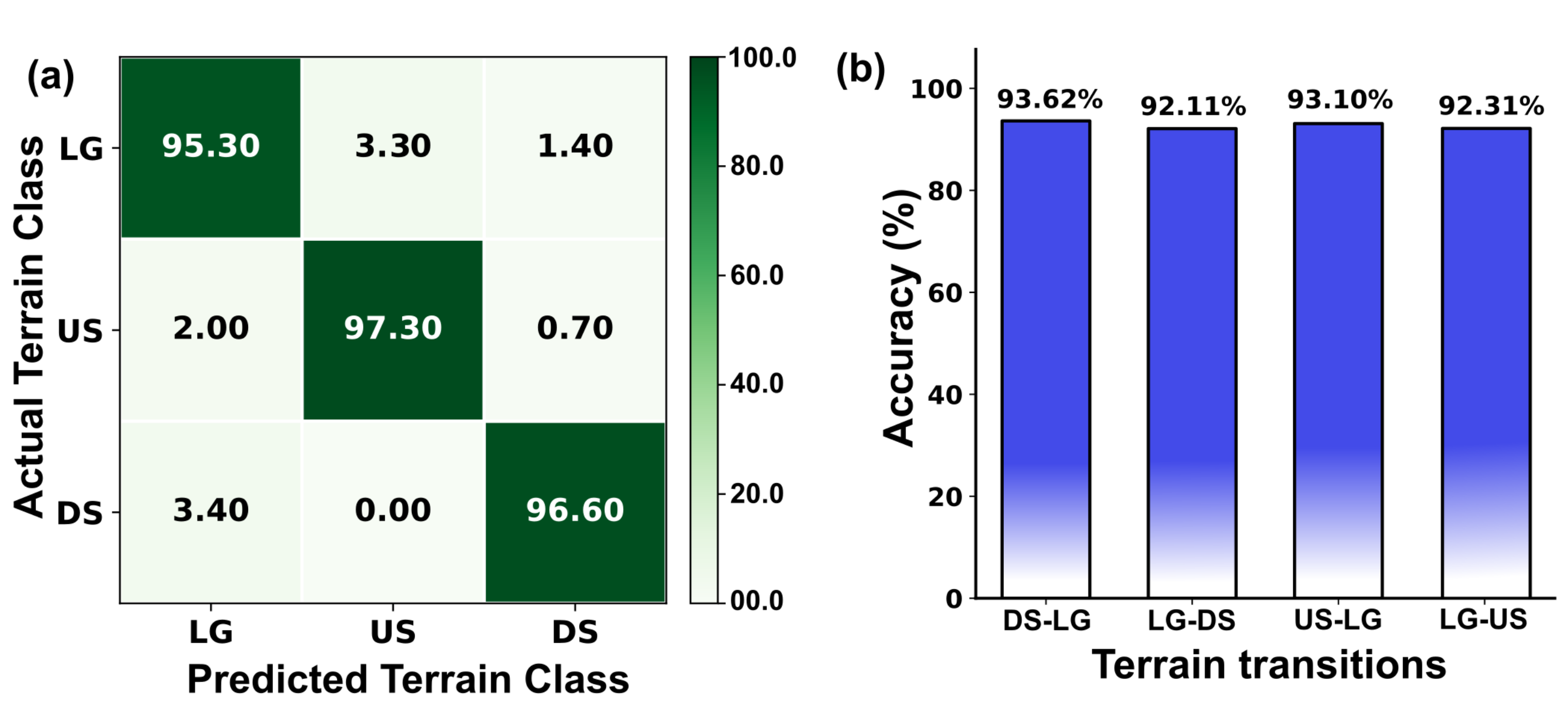}
\caption{ATP performance on the held-out test set. (a) Confusion matrix.
(b) Accuracy within 200 ms of terrain transitions.}
\label{fig4}
\end{figure}

On the held-out test set, ATP achieves 96.3\% accuracy, with class-wise F1 scores
of 0.949, 0.970, and 0.972 for level ground, stair ascent, and stair descent,
respectively (Table~\ref{tab_atp}). The confusion matrix (Fig.~\ref{fig4}a)
shows that stair ascent and descent are rarely confused and most of the classification errors occur
between level ground and stairs.

Accuracy within a window of 200 ms on either side of the terrain transitions is observed in the range of
92.1--93.6\% across the four transition cases  (Fig.~\ref{fig4}b).
The terrain transition anticipation horizon is observed to be
$1460.0\pm996.5$ ms, measured from the ATP terrain
transition to the corresponding locomotion-mode transition. The large standard deviation relative to the mean reflects substantial trial-to-trial variation, driven by differences in approach geometry and the point at which the new terrain becomes visually dominant. This anticipatory output is passed to TALMI and the downstream
context-aware estimators.

\section{Terrain-Aware Locomotion Mode Identification}
\label{sec:talmi}

TALMI identifies seven locomotion modes: standing (ST), fidgeting (FG), level
walking (LW), right turn (RT), left turn (LT), stair ascent (SA), and stair
descent (SD). Fidgeting denotes leg movement while remaining stationary, such as
weight shifting, foot repositioning, or stepping in place. It is treated
separately from standing and walking to avoid triggering assistance during
preparatory movement.

\begin{figure*}[!t]
\centering
\includegraphics[scale=1.0]{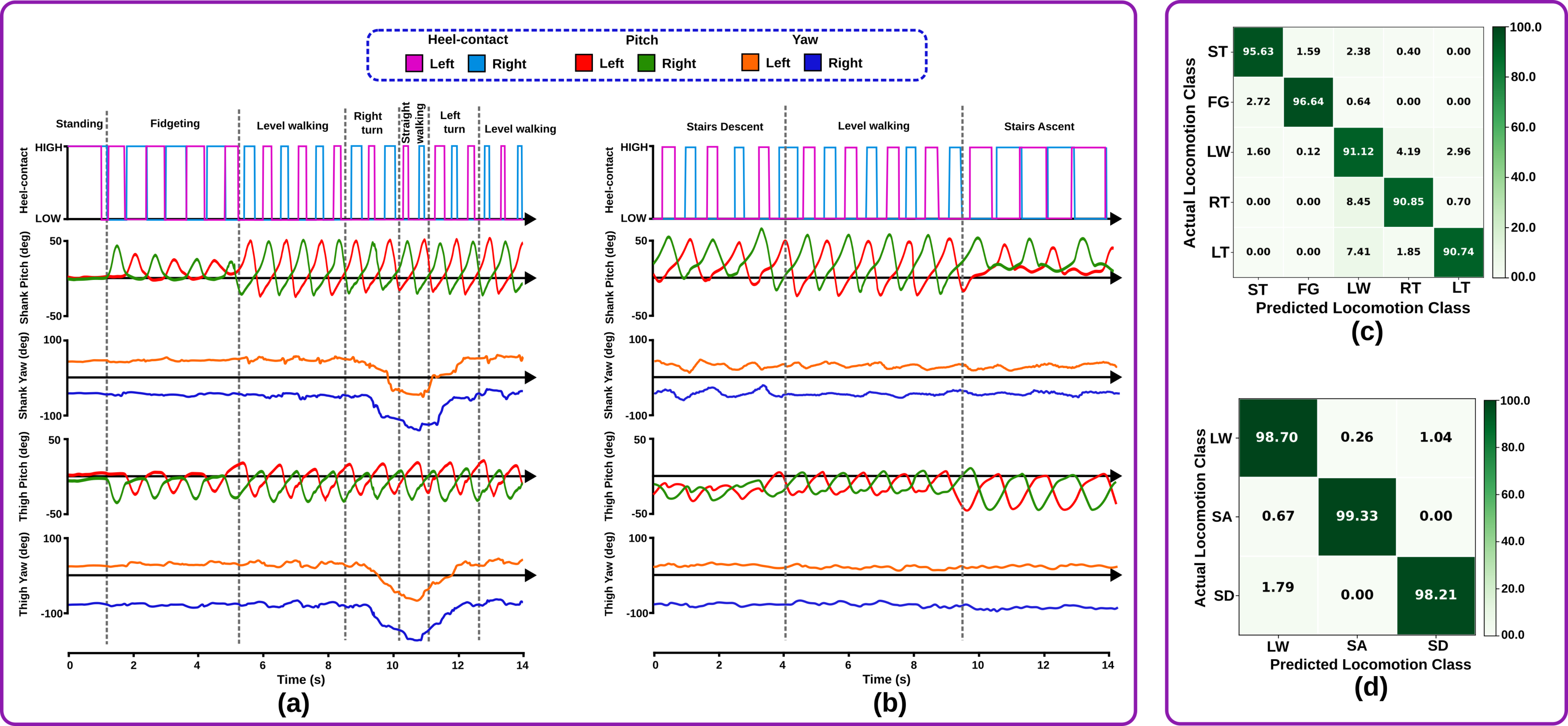}
\caption{Synchronized heel-contact and IMU traces used for annotation.
(a) Level-ground sequence. (b) stairs sequence. (c,d) TALMI confusion matrices.
FSR based heel-contact signals are used only for annotation.}
\label{fig5}
\end{figure*}

TALMI uses pitch and yaw from the four IMUs. Pitch captures sagittal-plane motion,
while yaw changes primarily during turns, enabling separation of turning from
straight walking. Differences in pitch amplitude and waveform further distinguish
stair ascent from descent. The synchronized heel-contact and IMU traces used for
annotation are shown in Fig.~\ref{fig5}.

Pitch and yaw being sampled at 100 Hz, a 30-sample (300 ms) window is used, advancing by
10 samples (100 ms), yielding a 10 Hz inference stream for locomotion mode estimation. For each IMU, seven pitch signal based temporal
statistical features (mean, RMS, maximum, minimum, range, net excursion, and standard
deviation) and three yaw based temporal statistical features (range, net excursion, and standard deviation)
are extracted. Absolute yaw is excluded because it depends on heading and the observed change
in yaw signals within the window captures turning motions. The resulting descriptor
$\mathbf{u}_t^{(i)}\in\mathbb{R}^{10}$ from each IMU is concatenated as:

\begin{equation}
\mathbf{u}_t=
\big[\mathbf{u}_t^{(1)\top},
\mathbf{u}_t^{(2)\top},
\mathbf{u}_t^{(3)\top},
\mathbf{u}_t^{(4)\top}\big]^{\top}
\in\mathbb{R}^{40}.
\label{eq:concat}
\end{equation}

The current descriptor and seven preceding descriptors form
$\mathbf{U}_t=[\mathbf{u}_{t-7},\ldots,\mathbf{u}_t]^\top\in\mathbb{R}^{8\times40}$,
corresponding to 1.0 s of temporal context.

\begin{figure*}[!t]
\centering
\includegraphics[scale=1.0]{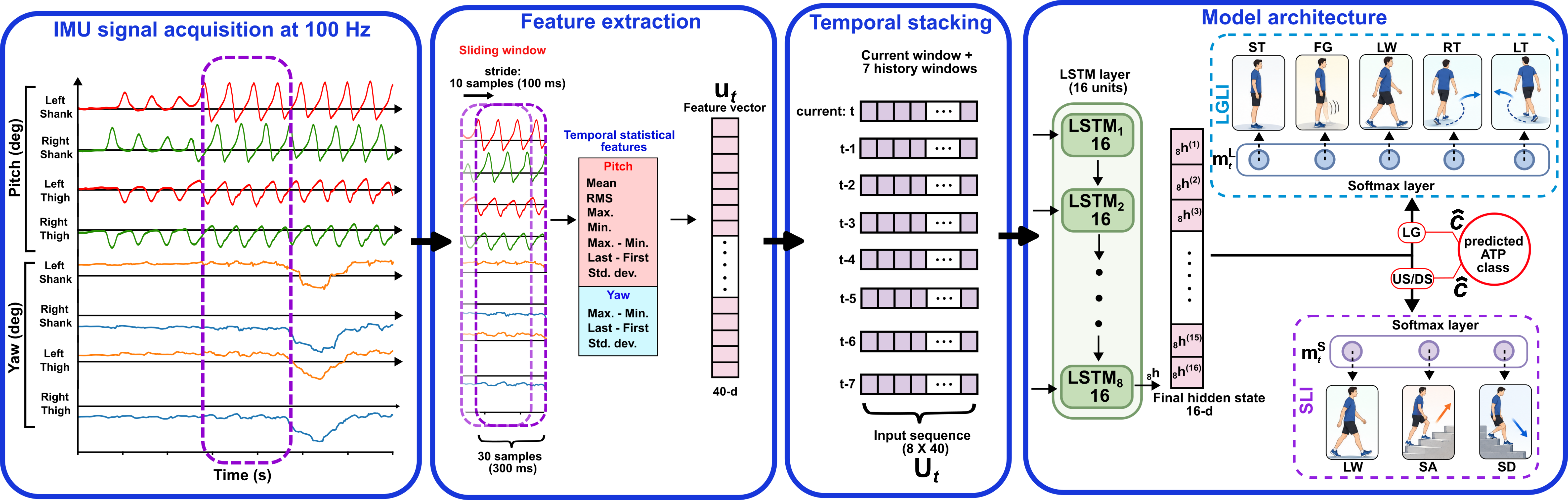}
\caption{TALMI pipeline from IMU acquisition and feature extraction to temporal
stacking and the LGLI and SLI softmax layers.}
\label{fig6}
\end{figure*}

TALMI uses two terrain-specific experts with identical architectures
(Fig.~\ref{fig6}). The Level Ground Locomotion Identification (LGLI) expert
classifies $\{\mathrm{ST},\mathrm{FG},\mathrm{LW},\mathrm{RT},\mathrm{LT}\}$,
whereas the Stairs Locomotion Identification (SLI) expert classifies
$\{\mathrm{LW},\mathrm{SA},\mathrm{SD}\}$. Each expert consists of a 16-unit LSTM
\cite{lstm} operating on the eight timesteps of $\mathbf{U}_t$, followed by a
softmax classifier. The SLI and LGLI are independent models and are trained individually. The ATP terrain
decision $\hat{c}$ selects the active expert:

\begin{equation}
\mathbf{m}_t=
\begin{cases}
\mathbf{m}^{\mathrm{L}}_t, & \hat{c}=\mathrm{LG}\\[1pt]
\mathbf{m}^{\mathrm{S}}_t, & \hat{c}\in\{\mathrm{US},\mathrm{DS}\}
\end{cases}
\label{eq:sel}
\end{equation}

\begin{equation}
\hat{\ell}_t=\arg\max_j[\mathbf{m}_t]_j
\label{eq:se2}
\end{equation}

This hierarchy restricts each expert to terrain-consistent classes. Both experts are trained for 50 epochs with
Adam optimizer, batch size 128, categorical cross-entropy, inverse-frequency class
weighting for further balancing the dataset, and a 70:30 training/validation split.

\begin{table}[!t]
\caption{TALMI CLASSIFICATION REPORTS FOR THE LGLI AND SLI EXPERTS}
\label{tab_talmi}
\centering
\footnotesize
\setlength{\tabcolsep}{4pt}
\begin{tabular}{|c|c|c|c|c|}
\hline
\textbf{Expert} & \textbf{Class} & \textbf{Precision} &
\textbf{Recall} & \textbf{F1 Score} \\
\hline
\multirow{6}{*}{\textbf{LGLI}}
& ST & 0.941 & 0.956 & 0.949 \\
\cline{2-5}
& FG & 0.985 & 0.966 & 0.976 \\
\cline{2-5}
& LW & 0.954 & 0.911 & 0.932 \\
\cline{2-5}
& RT & 0.772 & 0.908 & 0.835 \\
\cline{2-5}
& LT & 0.797 & 0.907 & 0.848 \\
\cline{2-5}
& \textbf{Accuracy} & \multicolumn{3}{c|}{0.937} \\
\hline
\multirow{4}{*}{\textbf{SLI}}
& LW & 0.982 & 0.987 & 0.984 \\
\cline{2-5}
& SA & 0.996 & 0.993 & 0.995 \\
\cline{2-5}
& SD & 0.986 & 0.982 & 0.984 \\
\cline{2-5}
& \textbf{Accuracy} & \multicolumn{3}{c|}{0.988} \\
\hline
\end{tabular}
\end{table}

LGLI and SLI achieve average accuracy of 93.7\% and 98.8\% (Table~\ref{tab_talmi}) respectively. LGLI errors
are concentrated between level walking and turns, with turn precisions of 0.772(right)
and 0.797(left), whereas SLI errors mainly occur between level walking and stairs.
Stair ascent and descent are rarely confused. Fidgeting achieves 0.985 precision,
with errors primarily assigned to standing.

\begin{figure*}[!t]
\centering
\includegraphics[scale=1.0]{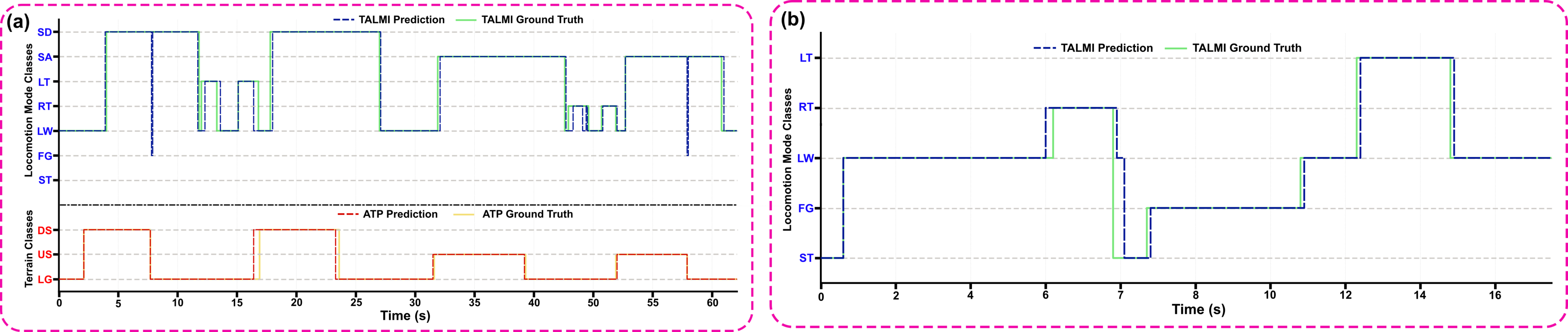}
\caption{Coupled ATP--TALMI inference for a participant excluded from training,
validation, and test partitions. (a) Mixed terrain route; (b) level-ground
segment containing LGLI classes.}
\label{fig7}
\end{figure*}

Fig.~\ref{fig7} shows coupled ATP--TALMI inference for an unseen participant, where the change in terrain output before the corresponding locomotion-mode transition confirms the anticipatory feature of the ATP.
A two-sample filter is applied to the 10 Hz locomotion mode inference stream, accepting a transition
only after two consecutive inferences agree. This suppresses isolated predictions
and adds at most 100 ms latency. Similar filtering has been used in IMU-only mode
recognition \cite{ma2025} where the transition context is provided upstream by
vision.

\begin{figure}[!t]
\centering
\includegraphics[scale=1.0]{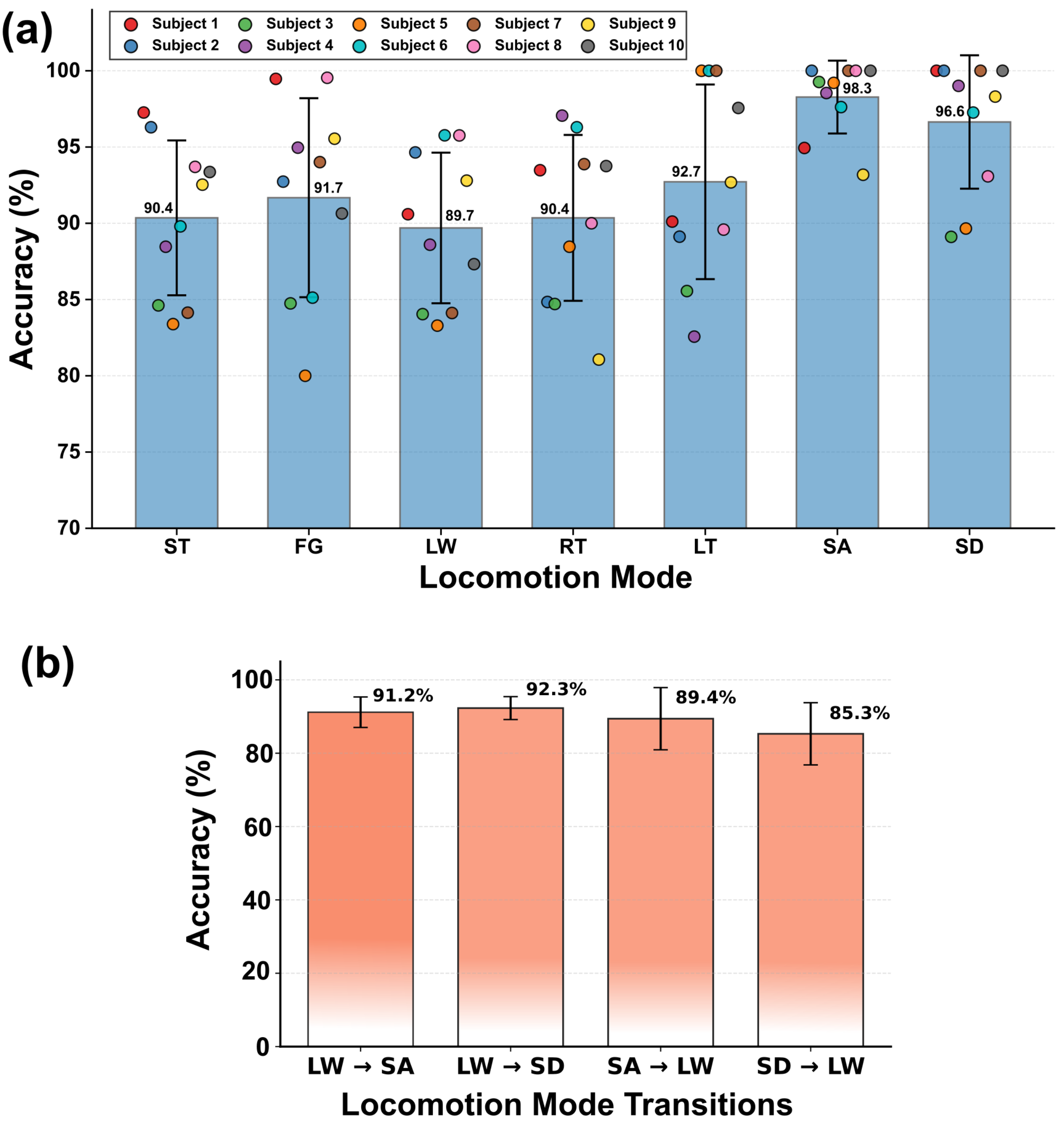}
\caption{Subject-independent TALMI evaluation. (a) Per-class recall under
leave-one-subject-out cross-validation. (b) Accuracy within 800 ms of mode
transitions on unseen participants.}
\label{fig8}
\end{figure}

TALMI transition detection latency is 171.4 $\pm$ 99.4 ms for late detection and
180.0 $\pm$ 130.3 ms for early detection, both with mean below 200 ms. Under leave-one-subject-out cross validation (Fig.\ref{fig8}(a)), mean recall is
near or above 90\% for all level-ground classes and above 96\% for both stair
classes while fidgeting remains above 91\%. Within a bidirectional window of 800 ms of true transitions on unseen
participants (Fig. \ref{fig8}(b)), accuracy is observed to be 85.3--92.3\%, with the lowest value for stair-descent-to-level-ground transitions.

\section{Locomotion-Context-Aware Estimation}
\label{sec:ctx}

\subsection{Probabilistic Locomotion Context and Input Formation}

Gait phase and walking speed depend on locomotion mode. In this work,instead of hard switching
between mode-specific estimators, TALMI's posterior is retained as a probabilistic
context input, allowing mode uncertainty to propagate to the downstream estimator.

The PLC vector $\boldsymbol{\gamma}_n\in\mathbb{R}^{3}$ is obtained from the active
TALMI expert as:

\begin{equation}
\boldsymbol{\gamma}_n=
\begin{cases}
\big[m^{\mathrm{S}}_{\mathrm{LW}},\,
m^{\mathrm{S}}_{\mathrm{SA}},\,
m^{\mathrm{S}}_{\mathrm{SD}}\big]^{\top},
& \hat{c}\in\{\mathrm{US},\mathrm{DS}\}\\[3pt]
\big[m^{\mathrm{L}}_{\mathrm{LW}},\,0,\,0\big]^{\top},
& \hat{c}=\mathrm{LG}
\end{cases}
\label{eq:plc}
\end{equation}

where the stair-expert entries represent level walking, stair ascent, and stair
descent probabilities. For level ground, only the level-walking posterior is
propagated because the corresponding stair classes are excluded by ATP.

TALMI operates at 10 Hz and the downstream estimator at 50 Hz. The PLC vector is
therefore held for five consecutive 50-Hz windows:

\begin{equation}
\boldsymbol{\gamma}^{(50)}_n
=\boldsymbol{\gamma}_{\lfloor n/5\rfloor},
\qquad n=0,1,2,\ldots
\label{eq:zoh}
\end{equation}

Only pitch signals are used for phase and speed estimation. For each of the four IMUs,
the seven pitch statistics are computed over a 15-sample (150 ms) window advancing
by 2 samples (20 ms), yielding:
$\mathbf{v}_n\in\mathbb{R}^{28}$ at 50 Hz. The feature vector is:

\begin{equation}
\mathbf{x}_n=
\big[\mathbf{v}_n^{\top},\,
\boldsymbol{\gamma}^{(50)\top}_n\big]^{\top}
\in\mathbb{R}^{31}
\label{eq:input}
\end{equation}

and the preceding 43 vectors are stacked as:

\begin{equation}
\mathbf{X}_n=
\big[\mathbf{x}_{n-43},\ldots,\mathbf{x}_n\big]^{\top}
\in\mathbb{R}^{44\times31}
\label{eq:input}
\end{equation}

This provides approximately 1.01 s of temporal context and the inputs are standardized
using training-set statistics.

\begin{figure*}[!t]
\centering
\includegraphics[scale=1.0]{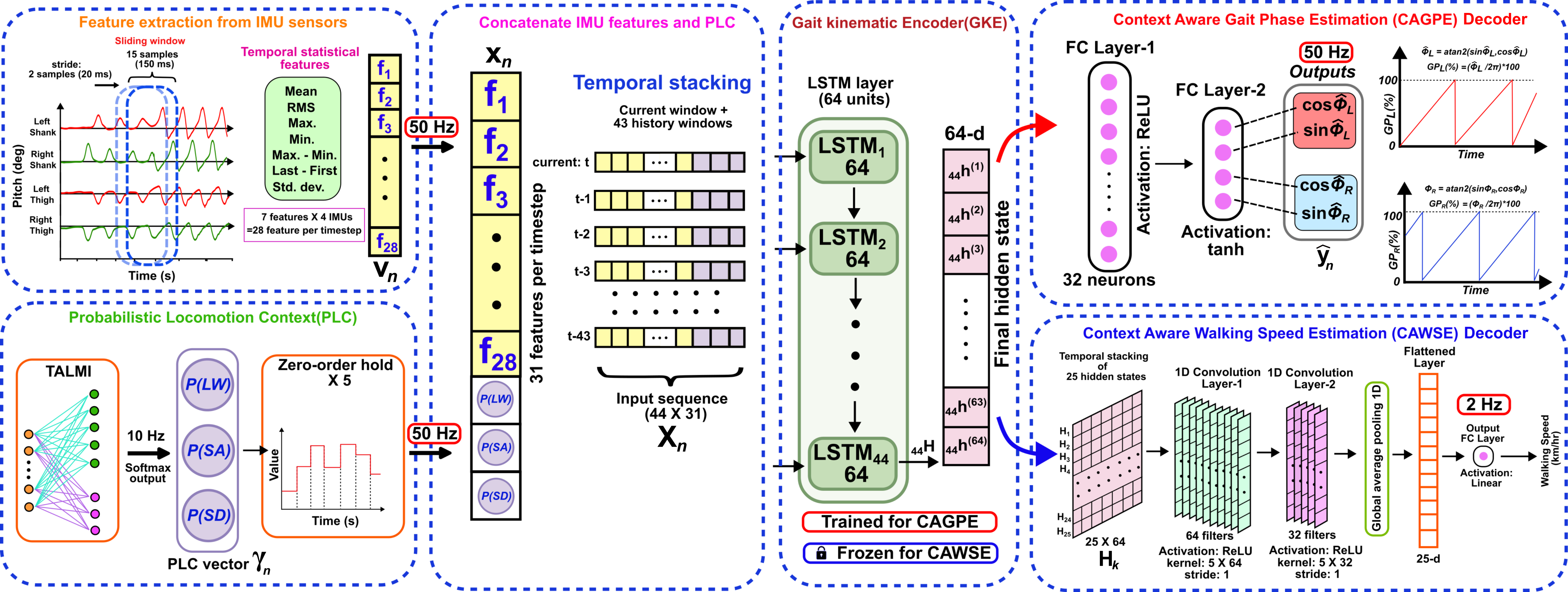}
\caption{CAGPE and CAWSE pipeline with PLC formation, the shared Gait Kinematic
Encoder, and the two decoder heads with their inference rates.}
\label{fig9}
\end{figure*}

\subsection{Shared Gait Kinematic Encoder}

The sequence $\mathbf{X}_n$ is processed by a 64-unit LSTM layer, the Gait Kinematic
Encoder (GKE), whose final hidden state
$\mathbf{h}_n\in\mathbb{R}^{64}$ is shared by both decoders (Fig.~\ref{fig9}).
The GKE is first trained jointly with the gait-phase decoder and then frozen while
the speed decoder is trained. This is essentially due to the fact that the gait phase estimation provides the more detailed temporal constraint throughout a gait cycle during training, whereas speed estimation is a much more aggregated effect. Thus, this allows the speed estimation decoder to be trained much conveniently without retraining the encoder.

\subsection{Context-Aware Gait Phase Estimation}

Gait phase is defined from 0\% at heel contact to 100\% at the next ipsilateral
heel contact. To avoid the discontinuity at cycle wrap related to gradient based optimizers, phase is represented on
the unit circle. For each limb, the decoder predicts $\cos\phi$ and $\sin\phi$:

\begin{equation}
\hat{\mathbf{y}}_n=
[\cos\hat{\phi}_L,\sin\hat{\phi}_L,
\cos\hat{\phi}_R,\sin\hat{\phi}_R]^{\top}
\end{equation}

The decoder comprises a 32-unit ReLU layer and a four-unit $\tanh$ output layer.
The phase angle and gait percentage are recovered as:

\begin{equation}
\hat{\phi}_{\{L,R\}}=
\operatorname{atan2}(\sin\hat{\phi},\cos\hat{\phi})
\bmod 2\pi,
\label{eq:phase}
\end{equation}

\begin{equation}
GP(\%)=\frac{\hat{\phi}}{2\pi}\times100
\label{eq:phase_gp}
\end{equation}

The loss combines output regression with a unit-circle constraint:

\begin{equation}
\mathcal{L}=
\frac{1}{N}\sum_n
\left\|\hat{\mathbf{y}}_n-\mathbf{y}_n\right\|^2
+\lambda\sum_{l\in\{L,R\}}
\left(\sin^2\hat{\phi}_l+\cos^2\hat{\phi}_l-1\right)^2
\label{eq:loss}
\end{equation}

with $\lambda=0.1$. The model is trained jointly with the GKE for 100 epochs using
Adam optimizer and batch size 32 and runs at 50 Hz.

\begin{figure}[!t]
\centering
\includegraphics[scale=1.0]{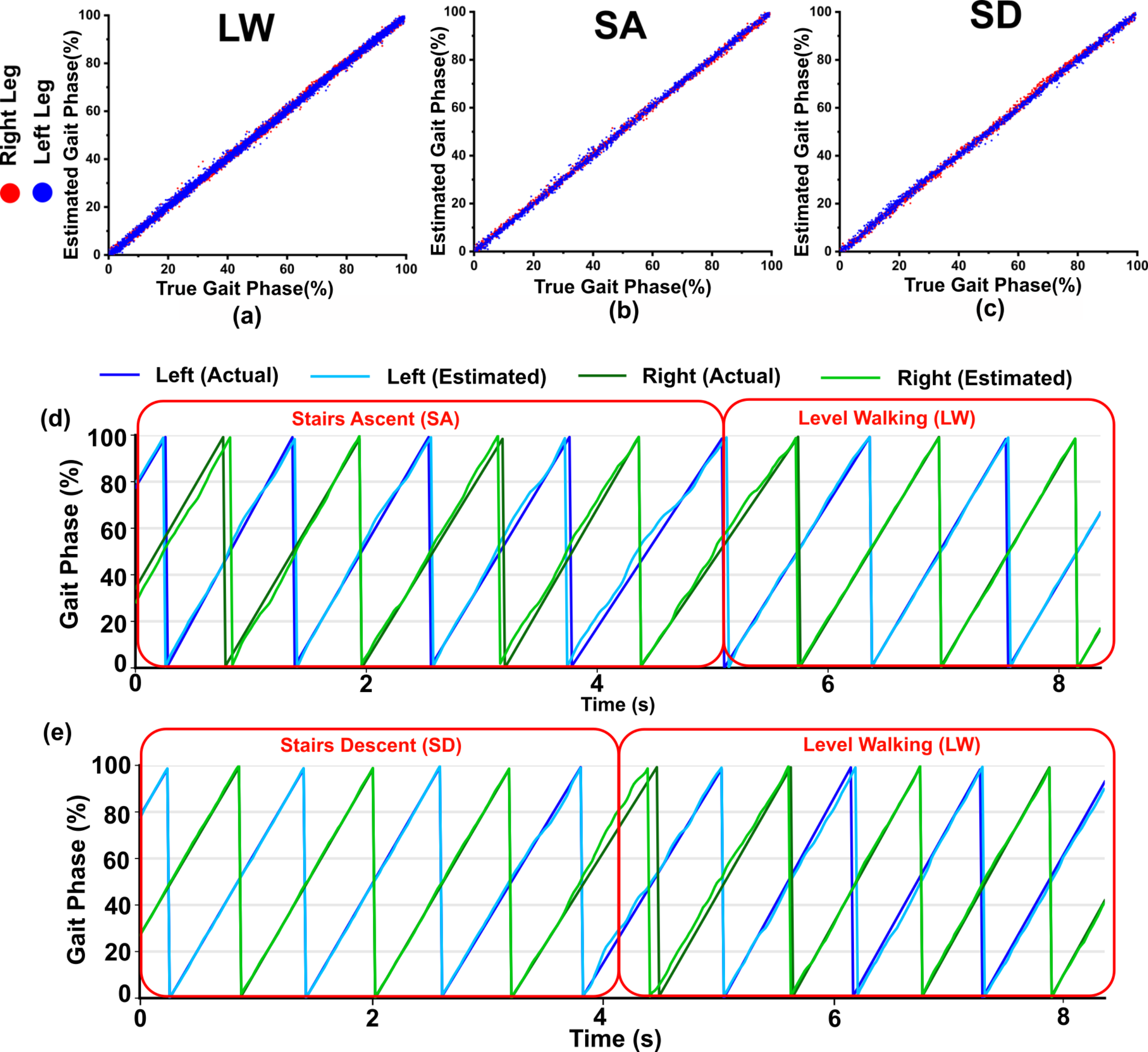}
\caption{CAGPE performance. (a)–(c) Estimated versus true gait phase for LW, SA, and SD. (d)–(e) Continuous bilateral phase estimation across stairs-to-level-walking transitions (stair ascent and stair descent, respectively).}
\label{fig10}
\end{figure}

Across LW, SA, and SD, the estimated phase follows the identity relationship over
the full 0--100\% range (Fig. \ref{fig10} (a)-(c)). LOSO RMSEs are 1.75\%, 2.97\% and
4.55\%, respectively (Fig.~\ref{fig11}(a)) for the gait phase estimation during the three locomotion modes. The larger error during stair descent
reflects greater subject-to-subject dispersion. The continuous prediction on an
unseen participant maintains bilateral phase through transitions (Fig. \ref{fig10} (d)-(e)), with small
deviations near mode changes that recover within one stride.

\subsection{Context-Aware Walking Speed Estimation}

CAWSE uses a short sequence of GKE states rather than a single hidden state.
Twenty-five consecutive states are stacked as:

\begin{equation}
\mathbf{H}_k=
[\mathbf{h}_{k-24},\ldots,\mathbf{h}_k]^{\top}
\in\mathbb{R}^{25\times64}
\end{equation}

The 25 states are non-overlapping, yielding one speed estimate every 25 encoder
outputs, or 2 Hz at the 50-Hz encoder rate. Two one-dimensional convolutional
layers with five-timestep filters and same padding process $\mathbf{H}_k$, followed
by global average pooling and a linear output predicting speed in km/h. The decoder is trained for 100 epochs using Adam optimizer, batch size 64, and mean-squared-error loss with the GKE frozen, yielding a train RMSE of 0.019 km/h.

\begin{figure}[!t]
\centering
\includegraphics[scale=1.0]{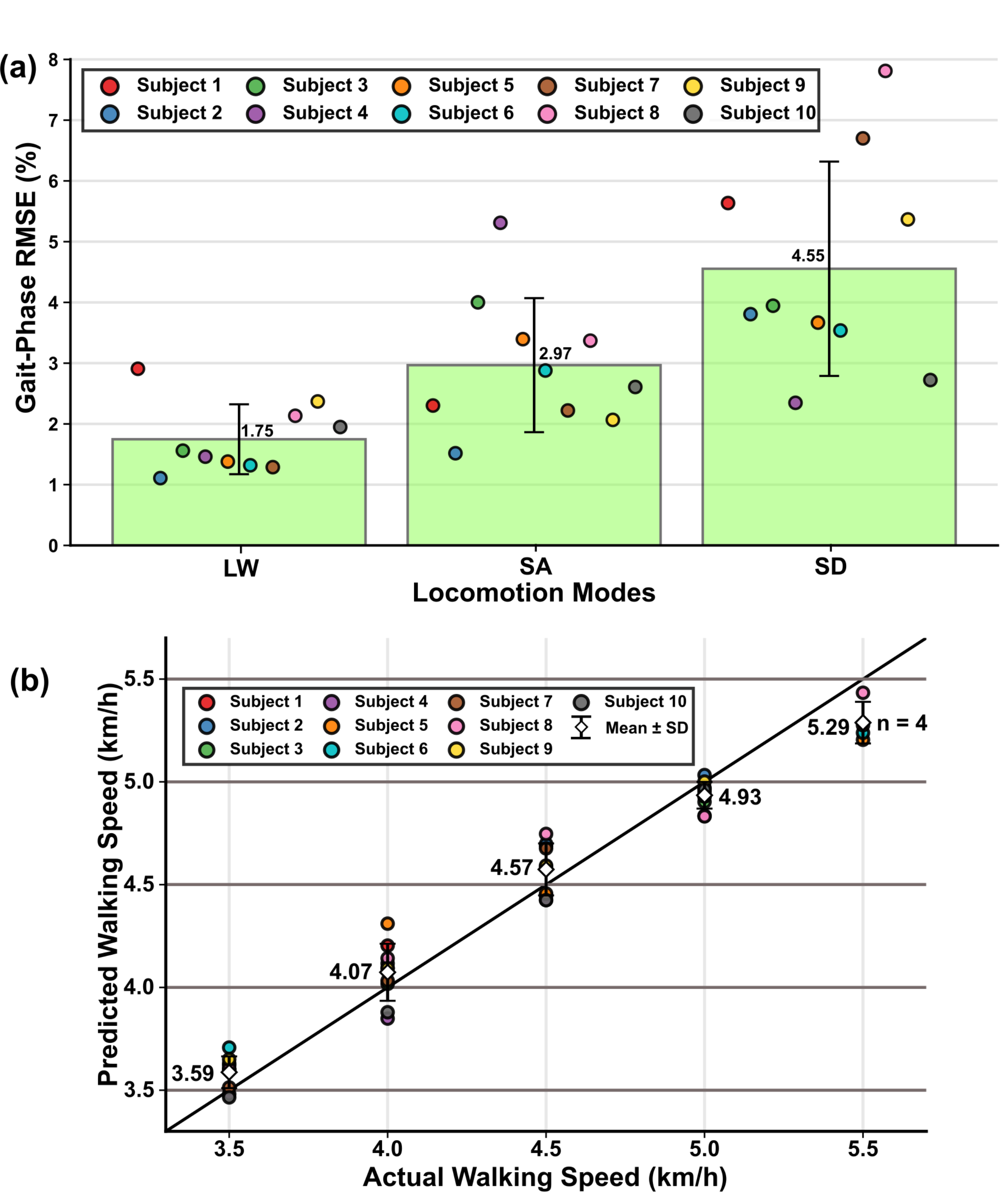}
\caption{LOSO evaluation of the context-aware estimators.
(a) CAGPE gait-phase RMSE for level walking, stair ascent, and stair descent;
(b) CAWSE predicted versus actual walking speed over 3.5--5.5~km/h}
\label{fig11}
\end{figure}

\begin{table}[!t]
\caption{TARGETED INFERENCE RATES AND MEASURED ONBOARD LATENCY}
\label{tab_rates}
\centering
\setlength{\tabcolsep}{5pt}
\begin{tabular}{|c|c|c|}
\hline
\textbf{Stage} & \textbf{Targeted Rate} & \textbf{Measured Latency (ms)}\\
\hline
ATP & 30 Hz & 13.57$\pm$1.22\\
\hline
TALMI & 10 Hz & 14.52$\pm$0.58\\
\hline
CAGPE & 50 Hz & 15.9$\pm$0.8\\
\hline
CAWSE & 2 Hz & 256.05$\pm$5.35\\
\hline
\end{tabular}
\end{table}

Under LOSO validation, speed predictions follow the identity relationship over 3.5–5.5 km/h (Fig. \ref{fig11}(b)), with a mean RMSE of 0.218 $\pm$ 0.072 km/h averaged across subjects and speeds, while the maximum observed deviation for subject means from ground truth is observed as 0.21 km/h. The shared GKE is trained for phase using a different
dataset and then frozen for CAWSE, demonstrating that the speed task can be added
without retraining the full pipeline. Table~\ref{tab_rates} summarizes the target
inference rates and onboard inference latency.

\section{Conclusion}
\label{sec:conclusion}

This paper presents a single visuo-inertial framework for estimating four
locomotion descriptors relevant to wearable-robot control: upcoming terrain,
locomotion mode, bilateral gait phase and walking speed. A chest-mounted RGB
camera provides anticipatory terrain information, while four lower-limb IMUs
provide inertial locomotion cues. These components are connected causally, with
the TALMI locomotion posterior passed as probabilistic context to a shared
Gait Kinematic Encoder and its phase and speed decoders.

ATP achieves 96.3\% terrain-classification accuracy and anticipates terrain
transitions by $1.46$ s on average. TALMI achieves 93.7\% and 98.8\% accuracy
for the level-ground and stairs experts respectively, with mean transition detection
latencies below 200 ms. The shared encoder yields leave-one-subject-out gait-phase
RMSEs of 1.75\%, 2.97\%, and 4.55\% for level walking, stair ascent, and stair
descent respectively, while speed estimation covers 3.5–5.5 km/h with a LOSO RMSE of 0.218 $\pm$ 0.072 km/h. Two design choices are central to the framework; probabilistic locomotion context rather than hard mode switching and a terrain-conditioned hierarchy that
combines datasets collected under different protocols. The inertial pathway
also uses gravity-referenced pitch and yaw Euler angles instead of raw inertial
channels, reducing dependence on sensor mounting. For qualitative validation, a supplementary video is provided demonstrating real-time inference of the complete pipeline, including terrain anticipation, locomotion-mode detection, bilateral gait-phase estimation and walking-speed estimation during indoor and outdoor locomotion.

Speed estimation was evaluated only for level walking using treadmill speed based annotated labels,
and terrain perception was limited to level ground and stairs. Future work will
extend speed estimation to stairs and overground walking, broaden terrain
coverage and evaluate closed-loop exoskeleton assistance using kinematic and
metabolic outcomes.

\balance

\end{document}